%% file: main.tex
\documentclass[journal,onecolumn]{IEEEtran}
\usepackage{blindtext}
\usepackage{graphicx}
\graphicspath{ {images/} }

\usepackage{amsthm}

\usepackage[export]{adjustbox}
\usepackage{cite}
\usepackage{amssymb}  % for math spacing
\usepackage{calc}
\usepackage[strings]{underscore}
\usepackage[cmex10]{amsmath}
\usepackage{mathrsfs}
\usepackage{cleveref}
\usepackage{array}
\newcolumntype{C}[1]{>{\centering\let\newline\\\arraybackslash\hspace{0pt}}m{#1}}
\newcolumntype{N}{@{}m{0pt}@{}}
\usepackage{mathtools}
\usepackage[font=footnotesize,subrefformat=parens,labelformat=parens]{subfig}
\usepackage{algorithm}
\usepackage[noend]{algpseudocode}
\usepackage{setspace}
\usepackage{color,soul}
\soulregister\cite7
\soulregister\ref7
\soulregister\pageref7

\makeatletter
\def\BState{\State\hskip-\ALG@thistlm}
\makeatother

\author{Akshay Shetty, University of Illinois Urbana-Champaign \\ Grace Xingxin Gao, Stanford University% <-this % stops a space
}
\usepackage{siunitx}

\begin{document}
\bstctlcite{IEEEexample:BSTcontrol}

\title{Trajectory Planning Under Stochastic and Bounded Sensing Uncertainties Using Reachability Analysis}

\maketitle
\thispagestyle{empty}
\textbf{Akshay Shetty} is a Ph.D. candidate in the Department of Aerospace Engineering at University of Illinois at Urbana-Champaign, and a visiting scholar at Stanford University. He received his M.S. degree in Aerospace Engineering from University of Illinois at Urbana-Champaign in 2017, and his B.Tech. degree in Aerospace Engineering from Indian Institute of Technology, Bombay, India in 2014. His research interests include safety, trajectory planning, and control for autonomous vehicles.

\textbf{Grace Xingxin Gao} Grace Xingxin Gao is an assistant professor in the Department of Aeronautics and Astronautics at Stanford University. Before joining Stanford University, she was an assistant professor at University of Illinois at Urbana-Champaign. She obtained her Ph.D. degree at Stanford University. Her research is on robust and secure positioning, navigation and timing with applications to manned and unmanned aerial vehicles, robotics, and power systems.
\vspace{0.5cm}
\begin{abstract}
\input{abstract}
\end{abstract}

\begin{IEEEkeywords}
\normalfont{\textbf{Trajectory Planning, Reachability Analysis, Probabilistic Zonotope, Global Navigation Satellite System (GNSS), Unmanned Aerial System (UAS)}}
\end{IEEEkeywords}

\IEEEpeerreviewmaketitle

\section{Introduction}
\input{introduction}

\section{Problem Formulation}
\input{problem_formulation}

\section{Set Representations and Operations}
\input{set_reps_ops}

\section{Reachability Analysis}
\input{reachability}

\section{Trajectory Planning}
\input{trajectory_planning}

\section{Simulations}
\input{simulations}

\section{Conclusions}
\input{conclusion_future}

\ifCLASSOPTIONcaptionsoff
  \newpage
\fi

\bibliographystyle{IEEEtran}
\bibliography{thesisrefs}

\iffalse
\begin{IEEEbiography}[{\includegraphics[width=1in,height=1.25in,clip,keepaspectratio]{Akshay_Shetty.png}}]{Akshay Shetty}
received the B.Tech. degree in aerospace engineering from Indian Institute of Technology, Bombay, India in 2014. He received the M.S. degree in aerospace engineering from University of Illinois at Urbana-Champaign in 2017. He is currently a PhD candidate at University of Illinois at Urbana-Champaign. His research interests include robotics, localization and path planning.
\end{IEEEbiography}

\begin{IEEEbiography}
[{\includegraphics[width=1in,height=1.25in,clip,keepaspectratio]{Grace_Gao.jpg}}]{Grace Xingxin Gao}
received the B.S. degree in mechanical engineering and the M.S. degree in electrical engineering from Tsinghua University, Beijing, China in 2001 and 2003. She received the PhD degree in electrical engineering from Stanford University in 2008. From 2008 to 2012, she was a research associate at Stanford University. Since 2012, she has been with University of Illinois at Urbana-Champaign, where she is presently an assistant professor in the Aerospace Engineering Department. Her research interests are systems, signals, control, and robotics.
\end{IEEEbiography}
\fi

\end{document}

%% file: abstract.tex
\label{sec:abstract}
Trajectory planning under uncertainty is an active research topic. Previous works predict state and state estimation uncertainties along trajectories to check for collision safety. They assume \textit{either} stochastic \textit{or} bounded sensing uncertainties. However, GNSS pseudoranges are typically modeled to contain stochastic uncertainties with additional biases in urban environments. Thus, given bounds for the bias, the planner needs to account for \textit{both} stochastic \textit{and} bounded sensing uncertainties. In our prior work \cite{shetty2020predicting} we presented a reachability analysis to predict state and state estimation uncertainties under stochastic and bounded uncertainties. However, we ignored the correlation between these uncertainties, leading to an imperfect approximation of the state uncertainty. In this paper we improve our reachability analysis by predicting state uncertainty as a function of independent quantities. We design a metric for the predicted uncertainty to compare candidate trajectories during planning. Finally, we validate the planner for GNSS-based urban navigation of fixed-wing UAS.

%% file: introduction.tex
\label{sec:introduction}

Recently there has been growing interest in autonomous applications of robotic systems for various purposes such as delivering goods, surveying areas of interest, and search and rescue \cite{zhou2016detecting,ke2016real,jwa2008information,zhou2014efficient,xu2016enhanced}. In general the problem of planning a trajectory for such systems has been extensively explored in literature. Traditional planning approaches proposed rapidly exploring random trees \cite{lavalle2006planning} and their variants \cite{karaman2011sampling}. However, these works do not account for uncertainty in the system state (position and orientation) that arises due to motion and sensing uncertainties.

Recent works \cite{van2011lqg,bry2011rapidly,claes2012collision,vaskov2019guaranteed} extended the traditional planners by accounting for motion and sensing uncertainties. These works predict the state uncertainty along candidate trajectories in order to find optimal, collision safe trajectories. In \cite{van2011lqg} the authors predict the state uncertainty for a system with linear-quadratic-Gaussian (LQG) control, while accounting for sensing uncertainties. The work in \cite{bry2011rapidly} predicts a distribution over the system states by considering the state estimation error distribution as well as all possible state estimates that could be realized in the future. In \cite{claes2012collision} the authors plan trajectories using a velocity obstacle paradigm in combination with a convex approximation of the sensing uncertainties. The work in \cite{vaskov2019guaranteed} predicts state uncertainty by using reachability analysis to account for tracking errors of a high-fidelity model in a lower-dimensional planning subspace. However, these works assume \textit{either} a stochastic \cite{van2011lqg,bry2011rapidly} \textit{or} a bounded \cite{claes2012collision,vaskov2019guaranteed} sensing uncertainty model, which may not always be valid.

For outdoor navigation, autonomous systems generally use Global Navigation Satellite System (GNSS) measurements for state estimation. GNSS pseudorange measurements are typically modeled to contain stochastic uncertainties along with an additional bias in urban environments due to signal reflections from nearby buildings. These effects are classified either as multipath, where both the direct and reflected signals from the same satellite are received; or as non-line-of-sight (NLOS), where only the reflected satellite signal is received \cite{misra2006global}. Generally, NLOS effects result in large biases in pseudorange measurements. Various outlier rejection techniques and three-dimensional (3D) map-based techniques have been proposed to detect and exclude the corresponding measurements \cite{peyraud2013non,wen2018exclusion,shetty2017covariance}. On the other hand, biases due to multipath effects are relatively lower. Previous works \cite{misra2006global,zhang2019new} have proposed methods to calculate the bounds for these multipath biases using a 3D map and the GNSS receiver architecture. Thus, given bounds for the additional bias, the trajectory planner needs to account for the presence of \textit{both} stochastic \textit{and} bounded sensing uncertainties.

In our prior work \cite{shetty2020predicting}, we used reachability analysis to predict state uncertainty under motion uncertainty and both stochastic and bounded sensing uncertainties. We recursively predicted the state uncertainty and the state estimation uncertainty, which are functions of the motion and sensing uncertainties along the trajectory. However, we ignored the correlation between the state and the state estimation uncertainties, which led to an imperfect approximation of the state uncertainty. Thus, in this paper we improve our prior reachability analysis and use it within a trajectory planning framework. This paper is based on our work in \cite{shetty2020trajectory}. The main contributions are listed as follows:
\begin{enumerate}
    \item We discuss the limitations in our prior reachability analysis \cite{shetty2020predicting} and present an improved analysis in order to predict state uncertainty as a function of independent (uncorrelated) quantities. We account for the presence of both stochastic and bounded sensing uncertainties in our analysis.
    
    \item Next, we combine our method to predict state uncertainty with an available trajectory planning framework \cite{ichter2017real}. In order to compare candidate trajectories within the planning framework, we design a metric for the size of the predicted state uncertainty.
    
    \item Finally, via simulations, we demonstrate the applicability of the trajectory planner for GNSS-based navigation of fixed-wing unmanned aerial systems (UAS) in an urban environment. Planning results are presented for two scenarios: a single UAS in a static environment, and multiple UAS in a shared airspace. We statistically validate the collision safety of the UAS by simulating multiple trajectories along the planned trajectories.
\end{enumerate}
The rest of the paper is organized as follows: we begin by formulating the problem in Section \ref{sec:prob-form}; in Section \ref{sec:set-reps-ops}, we present the set representations and operations used in the paper; in Section \ref{sec:reachability}, we mention the limitations of our prior work \cite{shetty2020predicting} and discuss our improved reachability analysis used to predict state uncertainty; Section \ref{sec:traj-planning} provides an overview of the trajectory planning framework \cite{ichter2017real} along with our metric for the size of the predicted state uncertainty; and in Section \ref{sec:simulations}, we demonstrate the planning results for urban GNSS-based navigation of a fixed-wing UAS.

%% file: problem_formulation.tex
\label{sec:prob-form}

We consider a non-linear discrete-time system with the following motion model:
\begin{equation}
    \label{eqn:motion_model}
    x_{k} = f(x_{k-1},u_{k-1}) + w_k,
\end{equation}
where $x_k$ is the state vector, $u_k$ is the input vector, $f$ is a non-linear function representing the kinematics, $w_k$ is the motion model error vector modeled as a zero-mean Gaussian distribution $\mathcal{N}(0,Q)$, and $k$ represents the time instant. For measurements, the following non-linear sensing model is considered:
\begin{equation}
    \label{eqn:mm}
    z_k = h(x_k) + \nu_k,
\end{equation}
where $z_k$ is the measurement vector, $h$ is a non-linear function representing the measurements, and $\nu_k$ is the sensing model error vector modeled as a Gaussian distribution with an uncertain mean, i.e., $\mathcal{N}(b_k, R_k)$. Here the mean $b_k$ can take any value in a bounded set $\mathcal{B}_k$. Thus, $\nu_k$ captures the presence of both stochastic and bounded sensing uncertainties.

Similar to prior trajectory planning work \cite{bry2011rapidly}, we assume there exists a low-level \texttt{CONNECT} function that connects a trajectory between two states. Thus, given states $x^i$ and $x^j$, the \texttt{CONNECT} function provides us the following:
\begin{equation}
    \label{eqn:connect-function}
    \left ( \check{X}^{i,j}, \check{U}^{i,j}, \check{K}^{i,j} \right ) = \texttt{CONNECT}( x^i, x^j ),
\end{equation}
where $\check{X}^{i,j}$ is the set of nominal states $( \check{x}_{\tau_i}, \check{x}_{\tau_i+1}, \cdots , \check{x}_{\tau_j-1} )$ and $\check{U}^{i,j}$ is the set of nominal inputs $( \check{u}_{\tau_i}, \check{u}_{\tau_i+1}, \cdots , \check{u}_{\tau_j-1} )$. These nominal states and inputs satisfy the nominal motion model from Equation (\ref{eqn:motion_model}):
\begin{equation}
\begin{gathered}
    \label{eqn:xu-nom}
    \check{x}_k = f(\check{x}_{k-1},\check{u}_{k-1}) \ \forall \ k \in [\tau_i+1,\tau_j-1], \\
    x^i = \check{x}_{\tau_i}, \ x^j = f(\check{x}_{\tau_j-1},\check{u}_{\tau_j-1}),
\end{gathered}
\end{equation}
where $\check{K}^{i,j}$ is the set of stabilizing linear state feedback control gains $( \check{K}_{\tau_i}, \check{K}_{\tau_i+1}, \cdots , \check{K}_{\tau_j-1} )$. Given the feedback control gain and an on-line state estimate $\hat{x}_k$, the total control input during execution is of the form:
\begin{equation}
    \label{eqn:feedback}
    {u}_k = \check{u}_k - \check{K}_k( \hat{x}_k - \check{x}_k ).
\end{equation}
%>>>>> This should not be here!!
%The planning of such a nominal trajectory along with computing the corresponding sets $\check{X}, \check{U}, \check{K}$ has been widely addressed in literature \cite{thrun2005probabilistic,karaman2011sampling} and is beyond the scope of this paper. For example, the optimal nominal states and inputs for a Dubins model can be easily obtained in closed form \cite{lavalle2006planning} and the feedback control gains can be obtained using a locally optimal linear-quadratic regulator (LQR) design.

Let $\mathcal{X}^{\text{obs}}$ denote the set of states representing obstacles in the environment. Given an initial state $x^{\text{init}}$ and a goal state $x^{\text{goal}}$, the problem for the trajectory planner is defined as:
\begin{equation}
\begin{gathered}
    \label{eqn:objective}
    \min_{(\check{X}, \check{U}, \check{K})} \text{cost}(\check{X},\check{U}), \\
    \text{subject to:} \ \Pr( x_k \in \mathcal{X}^{\text{obs}} ) < \delta,
\end{gathered}
\end{equation}
where $\delta$ is a specified threshold for the probability of collision, and $(\check{X}, \check{U}, \check{K})$ are concatenated sets representing the nominal trajectory from $x^{\text{init}}$ to $x^{\text{goal}}$ as follows:
\begin{equation}
    \label{eqn:concatenated}
    (\check{X}, \check{U}, \check{K}) = \left ( \texttt{CONNECT}(x^{\text{init}},x^1), \texttt{CONNECT}(x^{1},x^2), \cdots, \texttt{CONNECT}(x^{l},x^\text{goal}) \right ).
\end{equation}
For our trajectory planning problem, we set the cost function in Equation (\ref{eqn:objective}) to be the total length of the concatenated nominal trajectory from $x^{\text{init}}$ to $x^{\text{goal}}$.

%% file: set_reps_ops.tex
\label{sec:set-reps-ops}
\begin{figure}[b]
    \centering
  \subfloat[]{%
       \includegraphics[width=0.25\linewidth]{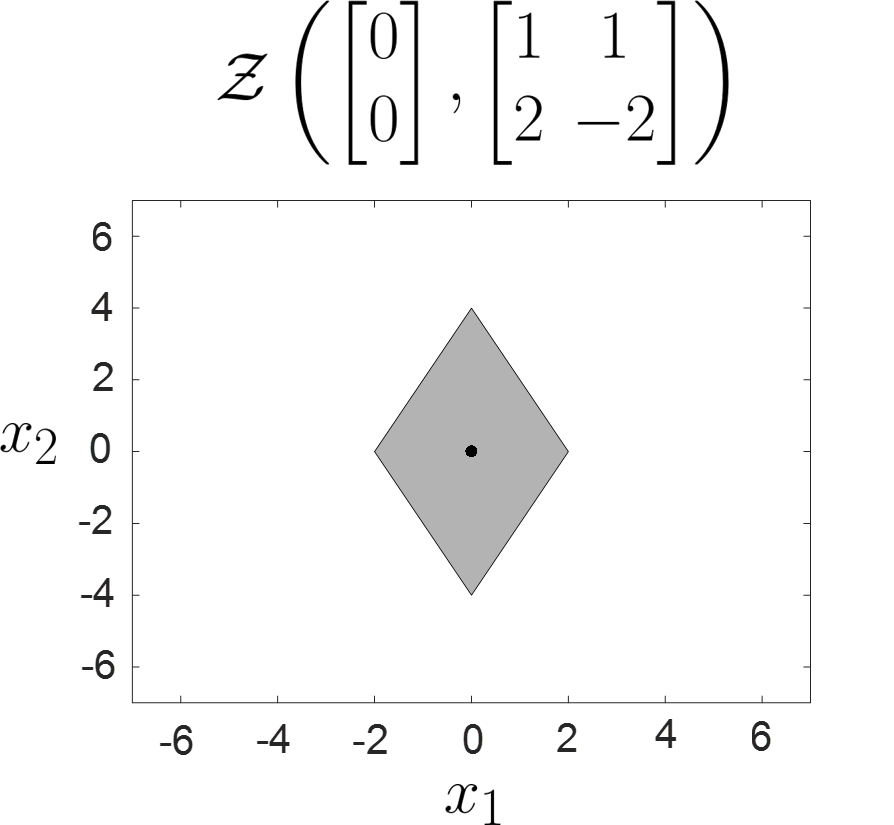}}
    \label{fig:set-vis-1}\hspace{1cm}
  \subfloat[]{%
        \includegraphics[width=0.25\linewidth]{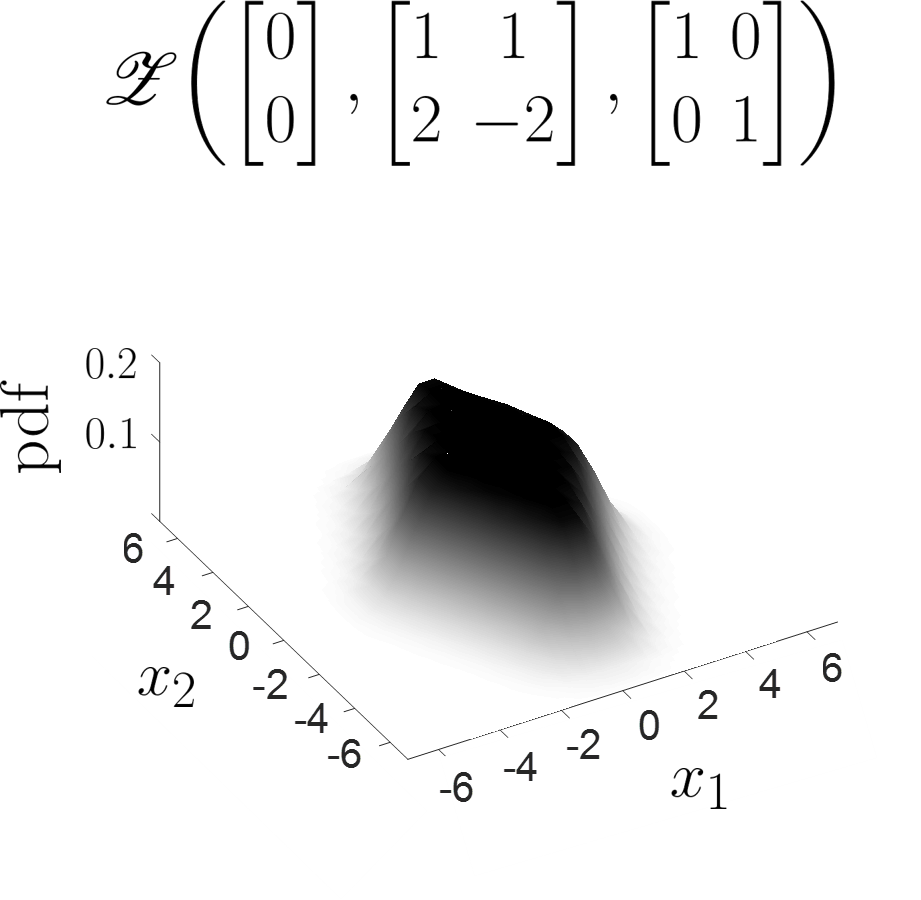}}
    \label{fig:set-vis-2}\hfill
  \caption{Example 2D visualizations of (a) a zonotope, and (b) a probabilistic zonotope.}
  \label{fig:set-vis}
\end{figure}

For our reachability analysis we use probabilistic zonotopes as the set representation since they are suitable to represent both stochastic and bounded sensing uncertainties \cite{althoff2009safety}. Probabilistic zonotopes have been shown to be computationally efficient and closed under linear transform and Minkowski sum operations \cite{althoff2010reachability}. A zonotope $\mathcal{P}$ is defined as follows:
\begin{equation}
    \label{eqn:zonotope}
    \mathcal{P} = \left \{ x \in \mathbb{R}^n | x = c_{\mathcal{P}} + \sum_{i=1}^{r} \beta_i \cdot g^{\{i\}}_{\mathcal{P}}, -1 \leq \beta_i \leq 1 \right \},
\end{equation}
where $c_\mathcal{P}$ is the center of the zonotope, and $g^{\{i\}}_{\mathcal{P}}$ are $n-$dimensional column vectors referred to as generators. The generators of a zonotope determine its shape relative to its center. The zonotope can be concisely written as $\mathcal{P} = \mathcal{Z}(c_\mathcal{P}, G_\mathcal{P})$, where $G_\mathcal{P} = \left [ g^{\{1\}}_{\mathcal{P}}, \dots , g^{\{r\}}_{\mathcal{P}} \right ]$ is the corresponding $n \times r$ generator matrix. Fig. \ref{fig:set-vis}(a) shows an example 2D zonotope along with its generator matrix. Zonotopes have been commonly used in literature to represent bounded uncertainties. In \cite{althoff2009safety}, the authors extended the representation to include stochastic uncertainties by defining a probabilistic zonotope as a Gaussian distribution with an uncertain and bounded mean:
\begin{equation}
\label{eqn:probabilistic-zonotope}
    \mathscr{P} = {\mathscr{Z}}(c_\mathscr{P}, G_\mathscr{P}, \Sigma_\mathscr{P}),
\end{equation}
where $c_\mathscr{P}$ and $G_\mathscr{P}$ represent the zonotope for the bounded uncertainty, and $\Sigma_\mathscr{P}$ is the Gaussian covariance representing the stochastic uncertainty. Fig. \ref{fig:set-vis}(b) provides an example visualization of a 2D probabilistic zonotope. Note that probabilistic zonotopes do not have a normalized distribution, and in fact can enclose multiple distributions.

The Minkowski sum operation between two probabilistic zonotopes is defined as \cite{althoff2009safety}:
\begin{equation}
\label{eqn:pz-minkowski-sum}
   \mathscr{P}_1 \oplus \mathscr{P}_2 = \mathscr{Z} \left ( c_{\mathscr{P}_1}+c_{\mathscr{P}_2}, [G_{\mathscr{P}_1}, G_{\mathscr{P}_2}], \Sigma_{\mathscr{P}_1} + \Sigma_{\mathscr{P}_2} \right ),
\end{equation}
whereas, the linear transform operation is defined as \cite{althoff2009safety}:
\begin{equation}
\label{eqn:pz-linear-transform}
T \cdot \mathscr{P} =  \mathscr{Z} \left ( T c_{\mathscr{P}}, T G_{\mathscr{P}}, T \Sigma_{\mathscr{P}} T^\top \right ).
\end{equation}
Note that the Minkowski sum operation in Equation (\ref{eqn:pz-minkowski-sum}) assumes $\mathscr{P}_1$ and $\mathscr{P}_2$ to be independent probabilistic zonotopes as it does not consider the correlation between the corresponding Gaussian covariances $\Sigma_{\mathscr{P}_1}$ and $\Sigma_{\mathscr{P}_2}$. This property gives rise to the limitation in our prior work and hence motivates our approach for computing stochastic reachable sets as discussed later in Section \ref{subsec:compute-reachability}.

%% file: reachability.tex
\label{sec:reachability}

\begin{figure}[t]
\centering
  \includegraphics[width=0.4\linewidth]{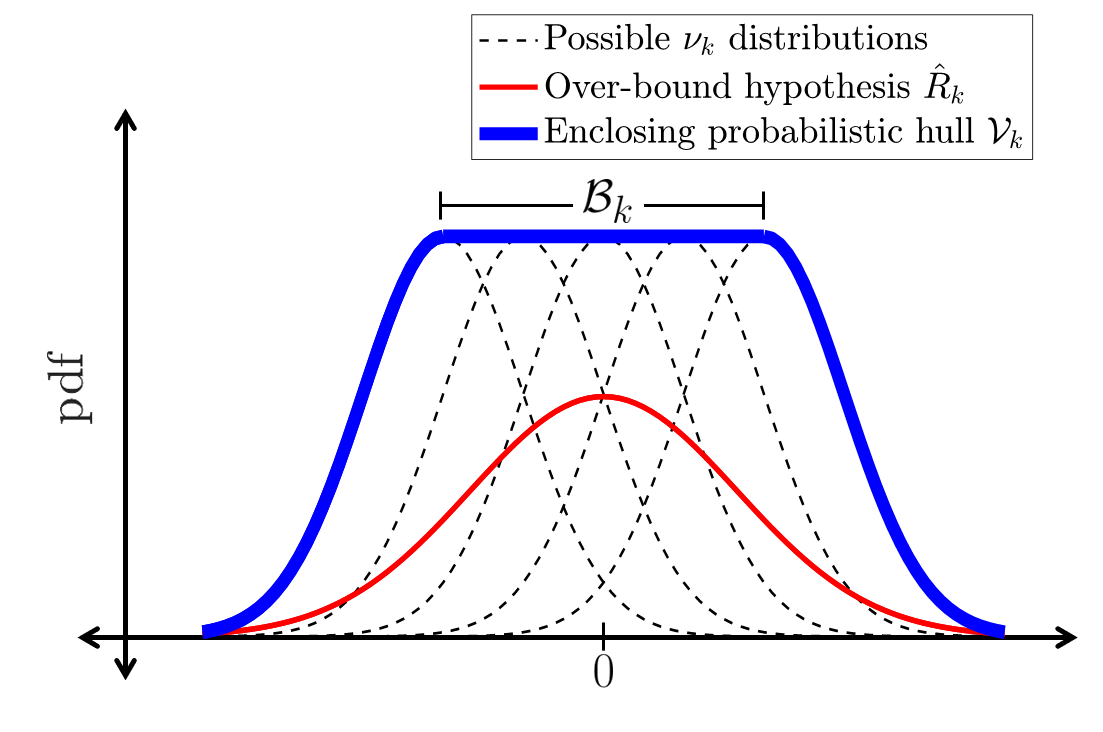}
  \caption{Visualization of over-bounding hypothesis for the state estimation filter used in the reachability analysis.}
  \label{fig:uncertainty}
\end{figure}

In this section, we present our reachability analysis used to predict the state uncertainty. We first briefly describe the state estimation filter used on-board. Next, we address the limitation in our prior reachability analysis, and derive the equations for computing the stochastic reachable sets as a function of independent quantities. Finally, we discuss the approximations we make to predict the state uncertainty.

\subsection{On-board State Estimation}
During trajectory execution, the true state $x_k$ will not be available to compute the control input. Thus, we need to estimate the state $\hat{x}_k$ in order to compute the total input as shown in Equation (\ref{eqn:feedback}). Given the non-linear motion and measurement models, we use an EKF as the state estimation filter. The prediction step of the EKF is performed as:
\begin{align}
    \label{eqn:ekf-pred-1}
    \bar{x}_k &= f(\hat{x}_{k-1},u_{k-1}), \\
    \label{eqn:ekf-pred-2}
    \bar{P}_k &= A_k P_{k-1} A_k^\top + Q,
\end{align}
where $A_k = \frac{\partial f}{\partial x}\Bigr|_{\substack{x=\check{x}_k}}$ and $Q$ is the covariance of the motion model uncertainty defined in Equation (\ref{eqn:motion_model}). For the EKF correction step, similar to our prior work, we choose an over-bounding hypothesis $\hat{R}_k$ as the measurement covariance matrix. The over-bounding is performed such that $\hat{R}_k$ matches the tail of the distribution at a desired confidence level. Fig. \ref{fig:uncertainty} illustrates the hypothesis for a simple one-dimensional case. Here, without loss of generality, we assume the sensing model error vector $\nu_k$ to be centered at origin. Note that our reachability analysis does not necessarily require choosing an over-bounding hypothesis for the EKF measurement covariance matrix. If desired, a different hypothesis can be chosen and used with the rest of the analysis. We choose the over-bounding hypothesis since it is equivalent to scaling or inflating the covariance matrix, which is a common approach for practical implementation of KF and its variants \cite{susi2017tuning,raanes2019adaptive,yang2006optimal}.

Once $\hat{R}_k$ has been computed, the EKF correction step is performed as:
\begin{align}
    \label{eqn:ekf-update-1}
    L_k &= \bar{P}_k C_k^\top (C_k \bar{P}_k C_k^\top + \hat{R}_k)^{-1}, \\
    \label{eqn:ekf-update-2}
    \hat{x}_k &= \bar{x}_k + L_k(z_k - h(\bar{x}_k)), \\
    \label{eqn:ekf-update-3}
    P_k &= \bar{P}_k - L_k C_k \bar{P}_k,
\end{align}
where $L$ is the Kalman gain and $C_k = \frac{\partial h}{\partial x}\Bigr|_{\substack{x=\check{x}_k}}$.

\subsection{Limitations in Our Prior Work}
\label{subsec:limitations-prior-work}

In our prior work we used the system equations (\ref{eqn:motion_model})-(\ref{eqn:feedback}) and state estimation equations (\ref{eqn:ekf-pred-1})-(\ref{eqn:ekf-update-3}) to derive the following recursive equations for the state $x_k$ and state estimation error $\tilde{x}_k$ ($= \hat{x}_k - x_k$):
\begin{align}
    \label{eqn:ITS-state}
    x_k &= (A_{k-1} - B_{k-1}\check{K}_{k-1})(x_{k-1} - \check{x}_{k-1}) - B_{k-1}\check{K}_{k-1}\tilde{x}_{k-1} + \check{x}_{k} + \mathcal{L}^f_{[s,\check{s}]_{k-1}} + w_k,\\ 
    \label{eqn:ITS-estimation-error}
    \tilde{x}_k = (I - L_kC_k) & A_{k-1}\tilde{x}_{k-1} + (I - L_kC_k)(\mathcal{L}^f_{[\hat{s},\check{s}]_{k-1}} - \mathcal{L}^f_{[{s},\check{s}]_{k-1}}) + L_k(\mathcal{L}^h_{[x,\check{x}]_k} - \mathcal{L}^h_{[\bar{x},\check{x}]_k}) - (I - L_kC_k)w_k + L_k\nu_k,
\end{align}
where $s^\top = [x^\top, u^\top]$ and $\mathcal{L}$ are Lagrange remainders resulting from the linearization of the non-linear models. For instance the remainder $\mathcal{L}^f_{[s,\check{s}]_{k-1}}$ is expressed as:
\begin{equation}
\label{eqn:lagrange1}
\begin{split}
    \mathcal{L}^{f_{(i)}}_{[s,\check{s}]_{k-1}} & = \frac{1}{2}(s_{k-1}-\check{s}_{k-1})^\top J^{f_{(i)}}_{s}(\xi)(s_{k-1}-\check{s}_{k-1}), \\
    \xi & \in \{ \check{s}_{k-1} + \eta (s_{k-1} - \check{s})_{k-1} | \ \eta \in [0,1] \},
\end{split}
\end{equation}
where subscript $(i)$ represents the $i^{th}$ element of the remainder vector and $J^{f_{(i)}}_{s}(\xi) = \frac{\partial^2 f_{(i)}(\xi)}{\partial s^2}$.

We then extended Equations (\ref{eqn:ITS-state}) and (\ref{eqn:ITS-estimation-error}) to set notations in order to recursively compute the stochastic reachable set $\mathcal{X}_k$ and the state estimation error set $\tilde{\mathcal{X}}_k$:
\begin{align}
    \label{eqn:ITS-state-set}
    \mathcal{X}_k &= (A_{k-1} - B_{k-1}\check{K}_{k-1})(\mathcal{X}_{k-1} - \check{x}_{k-1}) \oplus B_{k-1}\check{K}_{k-1}\tilde{\mathcal{X}}_{k-1} + \check{x}_{k} \oplus \mathcal{L}^f_{[s,\check{s}]_{k-1}} \oplus \mathcal{W}_k, \\
    \label{eqn:ITS-state-estimation-set}
    \tilde{\mathcal{X}}_k = (I \hspace{-0.1cm} - \hspace{-0.1cm} L_kC_k)&A_{k-1}\tilde{\mathcal{X}}_{k-1} \oplus (I \hspace{-0.1cm} - \hspace{-0.1cm} L_kC_k)(\mathcal{L}^f_{[\hat{s},\check{s}]_{k-1}} \hspace{-0.5cm} \oplus \mathcal{L}^f_{[s,\check{s}]_{k-1}}) \oplus L_k(\mathcal{L}^h_{[x,\check{x}]_k} \oplus \mathcal{L}^h_{[\bar{x},\check{x}]_k}) \oplus (I - L_kC_k)\mathcal{W}_k \oplus L_k\mathcal{V}_k,
\end{align}
where $\mathcal{W}_k$ and $\mathcal{V}_k$ are the set representations for uncertainties in the motion and sensing models, respectively. However, for the first Minkowski sum operation in Equation (\ref{eqn:ITS-state-set}) the sets $\mathcal{X}_{k-1}$ and $\tilde{\mathcal{X}}_{k-1}$ are not independent. By rewriting Equations (\ref{eqn:ITS-state-set}) and (\ref{eqn:ITS-state-estimation-set}) for the previous time instant, we observe that both $\mathcal{X}_{k-1}$ and $\tilde{\mathcal{X}}_{k-1}$ depend on $\mathcal{W}_{k-1}$. Thus, ignoring the correlation between $\mathcal{X}_{k-1}$ and $\tilde{\mathcal{X}}_{k-1}$ results in an imperfect approximation of the stochastic reachable set $\mathcal{X}_k$.

\subsection{Improved Computation of Stochastic Reachable Sets}
\label{subsec:compute-reachability}

In order to avoid the imperfect approximation in our prior work, we first derive equations for the state and the state estimation error as a function of independent (and hence uncorrelated) quantities. This is in contrast to Equation (\ref{eqn:ITS-state}) where $x_{k-1}$ and $\tilde{x}_{k-1}$ are correlated. We then extend the equations to set notations in order to compute the stochastic reachable set and the state estimation error set. Since the corresponding sets represent independent quantities, the Minkowski sum operations do not result in an imperfect approximation, in contrast to Equation (\ref{eqn:ITS-state-set}).

We prove using mathematical induction that the state $x_k$ and the state estimation error $\tilde{x}_k$ can be obtained as a function of independent quantities as follows:
\begin{equation}
    \label{eqn:state-independent-form}
    x_k = \check{x}_k + {}^1\!\phi_{
    k} (x_0 - \check{x}_0) + {}^2\!\phi_{k} \tilde{x}_0 + \sum_{n=1}^{k} {}^3\!\phi^{n}_{k} w_n + \sum_{n=1}^{k} {}^4\!\phi^{n}_{k} \nu_n + \sum_{n=0}^{k-1} {}^5\!\phi^{n}_{k} \mathcal{L}^f_{[s,\check{s}]_{n}} + \sum_{n=0}^{k-1} {}^6\!\phi^{n}_{k} \mathcal{L}^f_{[\hat{s},\check{s}]_{n}} + \sum_{n=0}^{k-1} {}^7\!\phi^{n}_{k} \mathcal{L}^h_{[x,\check{x}]_{n}} + \sum_{n=0}^{k-1} {}^8\!\phi^{n}_{k} \mathcal{L}^h_{[\bar{x},\check{x}]_{n}},
\end{equation}
\begin{equation}
    \label{eqn:estimation-independent-form}
    \tilde{x}_k = {}^2\!\tilde{\phi}_{k} \tilde{x}_0 + \sum_{n=1}^{k} {}^3\!\tilde{\phi}^{n}_{k} w_n + \sum_{n=1}^{k} {}^4\!\tilde{\phi}^{n}_{k} \nu_n + \sum_{n=0}^{k-1} {}^5\!\tilde{\phi}^{n}_{k} \mathcal{L}^f_{[s,\check{s}]_{n}} + \sum_{n=0}^{k-1} {}^6\!\tilde{\phi}^{n}_{k} \mathcal{L}^f_{[\hat{s},\check{s}]_{n}} + \sum_{n=0}^{k-1} {}^7\!\tilde{\phi}^{n}_{k} \mathcal{L}^h_{[x,\check{x}]_{n}} + \sum_{n=0}^{k-1} {}^8\!\tilde{\phi}^{n}_{k} \mathcal{L}^h_{[\bar{x},\check{x}]_{n}},
\end{equation}
where all $\phi_k$ and $\tilde{\phi}_k$ are matrix coefficients derived from system and state estimation matrices. Here $x_0$ is the initial state of the system, $\tilde{x_0}$ is the initial state estimation error, $w_n$ and $\nu_n$ are motion and sensing model errors independently sampled along the trajectory, and $\mathcal{L}$ are the Lagrange remainders.

\begin{proof}
For the induction base case, the initial state and state estimation error can be written in the above form as:
\begin{equation}
\begin{gathered}
    \label{eqn:induction-base-step}
    x_0 = \check{x}_0 + (x_0 - \check{x}_0), \\
    \tilde{x}_0 = \tilde{x}_0,
\end{gathered}    
\end{equation}
which can be obtained by setting ${}^1\!\phi_{0} = I$, ${}^i\!\phi_{0}  = O$ $\forall$ $i=\{2,\cdots,8\}$, ${}^2\!\tilde{\phi}_{0} = I$ in Equation (\ref{eqn:state-independent-form}), and ${}^i\!\tilde{\phi}_{0} = O$ $\forall$ $i=\{3,\cdots,8\}$ in Equation (\ref{eqn:estimation-independent-form}).

For the induction step, we first assume that Equations (\ref{eqn:state-independent-form}) and (\ref{eqn:estimation-independent-form}) hold true for time instant $k-1$, i.e.:
\begin{multline}
    \label{eqn:state-independent-form1}
    x_{k-1} = \check{x}_{k-1} + {}^1\!\phi_{k-1} (x_0 - \check{x}_0) + {}^2\!\phi_{k-1} \tilde{x}_0 + \sum_{n=1}^{k-1} {}^3\!\phi^{n}_{k-1} w_n + \sum_{n=1}^{k-1} {}^4\!\phi^{n}_{k-1} \nu_n + \sum_{n=0}^{k-2} {}^5\!\phi^{n}_{k-1} \mathcal{L}^f_{[s,\check{s}]_{n}} \\ + \sum_{n=0}^{k-2} {}^6\!\phi^{n}_{k-1} \mathcal{L}^f_{[\hat{s},\check{s}]_{n}} + \sum_{n=0}^{k-2} {}^7\!\phi^{n}_{k-1} \mathcal{L}^h_{[x,\check{x}]_{n}} + \sum_{n=0}^{k-2} {}^8\!\phi^{n}_{k-1} \mathcal{L}^h_{[\bar{x},\check{x}]_{n}},
\end{multline}
\begin{equation}
    \label{eqn:estimation-independent-form1}
    \tilde{x}_{k-1} = {}^2\!\tilde{\phi}_{k-1} \tilde{x}_0 + \sum_{n=1}^{k-1} {}^3\!\tilde{\phi}^{n}_{k-1} w_n + \sum_{n=1}^{k-1} {}^4\!\tilde{\phi}^{n}_{k-1} \nu_n + \sum_{n=0}^{k-2} {}^5\!\tilde{\phi}^{n}_{k-1} \mathcal{L}^f_{[s,\check{s}]_{n}} + \sum_{n=0}^{k-2} {}^6\!\tilde{\phi}^{n}_{k-1} \mathcal{L}^f_{[\hat{s},\check{s}]_{n}} + \sum_{n=0}^{k-2} {}^7\!\tilde{\phi}^{n}_{k-1} \mathcal{L}^h_{[x,\check{x}]_{n}} + \sum_{n=0}^{k-2} {}^8\!\tilde{\phi}^{n}_{k-1} \mathcal{L}^h_{[\bar{x},\check{x}]_{n}}.
\end{equation}
Next, in order to show that Equations (\ref{eqn:state-independent-form}) and (\ref{eqn:estimation-independent-form}) hold true for time instant $k$ we begin by replacing Equation (\ref{eqn:estimation-independent-form1}) in Equation (\ref{eqn:ITS-estimation-error}). This gives us the following expression for the state estimation error $\tilde{x}_k$:
\begin{align}
    {}& \hspace{-0.6cm}\tilde{x}_k = (I - L_kC_k)A_{k-1}{}^2\!\tilde{\phi}_{k-1}\tilde{x}_0 + (I - L_kC_k)A_{k-1}\sum_{n=1}^{k-1} {}^3\!\tilde{\phi}^{n}_{k-1} w_n - (I - L_kC_k)w_k + (I - L_kC_k)A_{k-1}\sum_{n=1}^{k-1} {}^4\!\tilde{\phi}^{n}_{k-1} \nu_n + L_k\nu_k \nonumber \\
    {}& \hspace{-0.2cm} + (I - L_kC_k)A_{k-1}\sum_{n=0}^{k-2} {}^5\!\tilde{\phi}^{n}_{k-1} \mathcal{L}^f_{[s,\check{s}]_{n}} - (I - L_kC_k)\mathcal{L}^f_{[{s},\check{s}]_{k-1}} + (I - L_kC_k)A_{k-1}\sum_{n=0}^{k-2} {}^6\!\tilde{\phi}^{n}_{k-1} \mathcal{L}^f_{[\hat{s},\check{s}]_{n}} + (I - L_kC_k)\mathcal{L}^f_{[\hat{s},\check{s}]_{k-1}} \nonumber \\
    {}& \hspace{-0.2cm} + (I - L_kC_k)A_{k-1}\sum_{n=0}^{k-2} {}^7\!\tilde{\phi}^{n}_{k-1} \mathcal{L}^h_{[x,\check{x}]_{n}} + L_k\mathcal{L}^h_{[x,\check{x}]_k} + (I - L_kC_k)A_{k-1}\sum_{n=0}^{k-2} {}^8\!\tilde{\phi}^{n}_{k-1} \mathcal{L}^h_{[\bar{x},\check{x}]_{n}} - L_k\mathcal{L}^h_{[\bar{x},\check{x}]_k}.
\end{align}
The above equation can then be written in the form of Equation (\ref{eqn:estimation-independent-form}) where the $\tilde{\phi}_{k}$ matrix coefficients can be obtained as:
\begin{align}
\begin{split}
    \label{eqn:matrix-coeffs-estimation}
    {}& {}^2\!\tilde{\phi}_{k} = (I - L_kC_k)A_{k-1}{}^2\!\tilde{\phi}_{k-1}, \\
    {}& {}^3\!\tilde{\phi}^{n}_{k} = (I - L_kC_k)A_{k-1}{}^3\!\tilde{\phi}^{n}_{k-1} \ \forall \ n \in \{1, \cdots, k-1\},\ {}^3\!\tilde{\phi}^{k}_{k} = -(I - L_kC_k), \\
    {}& {}^4\!\tilde{\phi}^{n}_{k} = (I - L_kC_k)A_{k-1}{}^4\!\tilde{\phi}^{n}_{k-1} \ \forall \ n \in \{1, \cdots, k-1\},\ {}^4\!\tilde{\phi}^{k}_{k} = L_k, \\
    {}& {}^5\!\tilde{\phi}^{n}_{k} = (I - L_kC_k)A_{k-1}{}^5\!\tilde{\phi}^{n}_{k-1} \ \forall \ n \in \{0, \cdots, k-2\},\ {}^5\!\tilde{\phi}^{k-1}_{k} = -(I - L_kC_k), \\
    {}& {}^6\!\tilde{\phi}^{n}_{k} = (I - L_kC_k)A_{k-1}{}^6\!\tilde{\phi}^{n}_{k-1} \ \forall \ n \in \{0, \cdots, k-2\},\ {}^6\!\tilde{\phi}^{k-1}_{k} = (I - L_kC_k), \\
    {}& {}^7\!\tilde{\phi}^{n}_{k} = (I - L_kC_k)A_{k-1}{}^7\!\tilde{\phi}^{n}_{k-1} \ \forall \ n \in \{0, \cdots, k-2\},\ {}^7\!\tilde{\phi}^{k-1}_{k} = L_k, \\
    {}& {}^8\!\tilde{\phi}^{n}_{k} = (I - L_kC_k)A_{k-1}{}^8\!\tilde{\phi}^{n}_{k-1} \ \forall \ n \in \{0, \cdots, k-2\},\ {}^8\!\tilde{\phi}^{k-1}_{k} = -L_k.
\end{split}
\end{align}
Similarly for state $x_k$, replacing Equations (\ref{eqn:state-independent-form1}) and (\ref{eqn:estimation-independent-form1}) in Equation (\ref{eqn:ITS-state}), we get the following expression:
\begin{align}
    {}& \hspace{-0.6cm} x_k = \check{x}_k\! +\! (A_{k-1}\! -\! B_{k-1}\check{K}_{k-1}){}^1\!\phi_{k-1} (x_0\! -\! \check{x}_0)\! +\! \left ( (A_{k-1}\! -\! B_{k-1}\check{K}_{k-1}){}^2\!\phi_{k-1}\! -\!B_{k-1}\check{K}_{k-1}{}^2\!\tilde{\phi}_{k-1}  \right ) \tilde{x}_0\! +\! \nonumber \\
    {}& \hspace{-0.6cm} \sum_{n=1}^{k-1} \left ( (A_{k-1}\! -\! B_{k-1}\check{K}_{k-1}){}^3\!\phi^{n}_{k-1}\! -\!B_{k-1}\check{K}_{k-1}{}^3\!\tilde{\phi}^{n}_{k-1}\right )w_n\! +\! w_k\! +\! \sum_{n=1}^{k-1} \left ( (A_{k-1}\! -\! B_{k-1}\check{K}_{k-1}){}^4\!\phi^{n}_{k-1}\! -\!B_{k-1}\check{K}_{k-1}{}^4\!\tilde{\phi}^{n}_{k-1}\right )v_n\! +\! \nonumber \\
    {}& \hspace{-0.6cm} \sum_{n=0}^{k-2} \left ( (A_{k-1}\! -\! B_{k-1}\check{K}_{k-1}){}^5\!\phi^{n}_{k-1}\! -\!B_{k-1}\check{K}_{k-1}{}^5\!\tilde{\phi}^{n}_{k-1} \right ) \mathcal{L}^f_{[s,\check{s}]_{n}}\! +\! \mathcal{L}^f_{[s,\check{s}]_{k-1}}\! +\! \sum_{n=0}^{k-2} \left ( (A_{k-1}\! -\! B_{k-1}\check{K}_{k-1}){}^6\!\phi^{n}_{k-1}\! -\!B_{k-1}\check{K}_{k-1}{}^6\!\tilde{\phi}^{n}_{k-1} \right ) \mathcal{L}^f_{[\hat{s},\check{s}]_{n}}\! \nonumber \\
    {}& \hspace{-0.6cm} +\! \sum_{n=0}^{k-2} \left ( (A_{k-1}\! -\! B_{k-1}\check{K}_{k-1}){}^6\!\phi^{n}_{k-1}\! -\!B_{k-1}\check{K}_{k-1}{}^6\!\tilde{\phi}^{n}_{k-1} \right ) \mathcal{L}^h_{[x,\check{x}]_{n}}\! +\! \sum_{n=0}^{k-2} \left ( (A_{k-1}\! -\! B_{k-1}\check{K}_{k-1}){}^7\!\phi^{n}_{k-1}\! -\!B_{k-1}\check{K}_{k-1}{}^7\!\tilde{\phi}^{n}_{k-1} \right ) \mathcal{L}^h_{[\bar{x},\check{x}]_{n}}.
\end{align}
The above equation can then be written in the form of Equation (\ref{eqn:state-independent-form}) where the ${\phi}_{k}$ matrix coefficients can be obtained as:
\begin{align}
\begin{split}
    \label{eqn:matrix-coeffs-state}
    {}& {}^1\!\phi_{k} = (A_{k-1} - B_{k-1}\check{K}_{k-1}){}^1\!\phi_{k-1}, \\
    {}& {}^2\!\phi_{k} = (A_{k-1} - B_{k-1}\check{K}_{k-1}){}^2\!\phi_{k-1} -B_{k-1}\check{K}_{k-1}{}^2\!\tilde{\phi}_{k-1}, \\
    {}& {}^3\!\phi^{n}_{k} = (A_{k-1} - B_{k-1}\check{K}_{k-1}){}^3\!\phi^{n}_{k-1} -B_{k-1}\check{K}_{k-1}{}^3\!\tilde{\phi}^{n}_{k-1} \ \forall \ n \in \{1, \cdots, k-1\}, \ {}^3\!\phi^{k}_{k} = I, \\
    {}& {}^4\!\phi^{n}_{k} = (A_{k-1} - B_{k-1}\check{K}_{k-1}){}^4\!\phi^{n}_{k-1} -B_{k-1}\check{K}_{k-1}{}^4\!\tilde{\phi}^{n}_{k-1} \ \forall \ n \in \{1, \cdots, k-1\}, \ {}^4\!\phi^{k}_{k} = O, \\ 
    {}& {}^5\!\phi^{n}_{k} = (A_{k-1} - B_{k-1}\check{K}_{k-1}){}^5\!\phi^{n}_{k-1} -B_{k-1}\check{K}_{k-1}{}^5\!\tilde{\phi}^{n}_{k-1} \ \forall \ n \in \{0, \cdots, k-2\}, \ {}^5\!\phi^{k-1}_{k} = I, \\
    {}& {}^6\!\phi^{n}_{k} = (A_{k-1} - B_{k-1}\check{K}_{k-1}){}^6\!\phi^{n}_{k-1} -B_{k-1}\check{K}_{k-1}{}^6\!\tilde{\phi}^{n}_{k-1} \ \forall \ n \in \{0, \cdots, k-2\}, \ {}^6\!\phi^{k-1}_{k} = O, \\
    {}& {}^7\!\phi^{n}_{k} = (A_{k-1} - B_{k-1}\check{K}_{k-1}){}^7\!\phi^{n}_{k-1} -B_{k-1}\check{K}_{k-1}{}^7\!\tilde{\phi}^{n}_{k-1} \ \forall \ n \in \{0, \cdots, k-2\}, \ {}^7\!\phi^{k-1}_{k} = O,\\ 
    {}& {}^8\!\phi^{n}_{k} = (A_{k-1} - B_{k-1}\check{K}_{k-1}){}^8\!\phi^{n}_{k-1} -B_{k-1}\check{K}_{k-1}{}^8\!\tilde{\phi}^{n}_{k-1} \ \forall \ n \in \{0, \cdots, k-2\}, \ {}^8\!\phi^{k-1}_{k} = O.
    \end{split}
\end{align}
Thus, by the principle of induction, we show that Equations (\ref{eqn:state-independent-form}) and (\ref{eqn:estimation-independent-form}) hold true for all $k \in \{0, 1, \cdots\}$. The matrix coefficients $\phi_k$ and $\tilde{\phi}_k$ are obtained recursively using Equations (\ref{eqn:matrix-coeffs-estimation}) and (\ref{eqn:matrix-coeffs-state}).
\end{proof}

Once the above equations for the state and state estimation errors have been derived, we extend them to set notations in order to compute the stochastic reachable set $\mathcal{X}_k$ and the state estimation error set $\tilde{\mathcal{X}}_k$ as follows:
\begin{equation}
    \label{eqn:stochastic-reachable-set}
    \mathcal{X}_k = \check{x}_k \oplus {}^1\!\phi_{k} (\mathcal{X}_0 - \check{x}_0) \oplus {}^2\!\phi_{k} \tilde{\mathcal{X}}_0 \oplus \sum_{n=1}^{k} {}^3\!\phi^{n}_{k} \mathcal{W}_n \oplus \sum_{n=1}^{k} {}^4\!\phi^{n}_{k} \mathcal{V}_n \oplus \sum_{n=0}^{k-1} {}^5\!\phi^{n}_{k} \mathcal{L}^f_{[s,\check{s}]_{n}} \oplus \sum_{n=0}^{k-1} {}^6\!\phi^{n}_{k} \mathcal{L}^f_{[\hat{s},\check{s}]_{n}} \oplus \sum_{n=0}^{k-1} {}^7\!\phi^{n}_{k} \mathcal{L}^h_{[x,\check{x}]_{n}} \oplus \sum_{n=0}^{k-1} {}^8\!\phi^{n}_{k} \mathcal{L}^h_{[\bar{x},\check{x}]_{n}},
\end{equation}
\begin{equation}
    \label{eqn:state-estimation-error-set}
    \tilde{\mathcal{X}}_k = {}^2\!\tilde{\phi}_{k} \tilde{\mathcal{X}}_0 + \sum_{n=1}^{k} {}^3\!\tilde{\phi}^{n}_{k} \mathcal{W}_n + \sum_{n=1}^{k} {}^4\!\tilde{\phi}^{n}_{k} \mathcal{V}_n + \sum_{n=0}^{k-1} {}^5\!\tilde{\phi}^{n}_{k} \mathcal{L}^f_{[s,\check{s}]_{n}} + \sum_{n=0}^{k-1} {}^6\!\tilde{\phi}^{n}_{k} \mathcal{L}^f_{[\hat{s},\check{s}]_{n}} + \sum_{n=0}^{k-1} {}^7\!\tilde{\phi}^{n}_{k} \mathcal{L}^h_{[x,\check{x}]_{n}} + \sum_{n=0}^{k-1} {}^8\!\tilde{\phi}^{n}_{k} \mathcal{L}^h_{[\bar{x},\check{x}]_{n}}.
\end{equation}
Here the sets $\mathcal{X}_0$, $\tilde{\mathcal{X}}_0$, $\mathcal{W}_n$, and $\mathcal{V}_n$ represent independent quantities and hence the Minkowski sum operations do not result in an imperfect approximation of the stochastic reachable sets.

\subsection{Predicting State Uncertainty}
\label{subsec:predict-state-uncertainty}
Once we have an expression to compute the stochastic reachable sets, we perform the following two approximations in order to predict the state uncertainty:
\begin{enumerate}
    \item We heuristically bound the number of Minkowski sum operations in Equation (\ref{eqn:stochastic-reachable-set}). From Equations (\ref{eqn:matrix-coeffs-estimation}) and (\ref{eqn:matrix-coeffs-state}) we observe that the matrix coefficients contain contractive terms, which consequently results in some quantities having a negligible contribution in the computation of $\mathcal{X}_k$. Thus, we only consider quantities whose corresponding matrix coefficients have a Frobenius norm $\left \| {}^i\!\phi^n_{k} \right \|_F$ higher than a specified threshold ${}^i\!\zeta$.
    \item We approximate the Lagrange remainders $\mathcal{L}$ with a Gaussian distribution $\widehat{\mathcal{L}}$, as presented in our prior work \cite{shetty2020predicting}.
\end{enumerate}
Thus, we obtain the following expression for the predicted state uncertainty $\underline{\mathcal{X}}_k$:
\begin{equation}
    \label{eqn:predicted-state-uncertainty}
    \underline{\mathcal{X}}_k = \check{x}_k  \oplus {}^1\!\phi_{k} (\mathcal{X}_0 - \check{x}_0) \oplus {}^2\!\phi_{k} {\tilde{\mathcal{X}}_0} \oplus \sum_{n \in {}^3\!\mathcal{N}_k} {}^3\!\phi^{n}_{k} {\mathcal{W}_n} \oplus \sum_{n \in {}^4\!\mathcal{N}_k} {}^4\!\phi^{n}_{k} {\mathcal{V}_n} \\ \oplus \sum_{n \in {}^5\!\mathcal{N}_k} {}^5\!\phi^{n}_{k} {{\widehat{\mathcal{L}}}^f_{1,n}} \oplus \sum_{n \in {}^6\!\mathcal{N}_k} {}^6\!\phi^{n}_{k} {\widehat{\mathcal{L}}^f_{2,n}} \oplus \sum_{n \in {}^7\!\mathcal{N}_k} {}^7\!\phi^{n}_{k} {\widehat{\mathcal{L}}^h_{1,n}} \oplus \sum_{n \in {}^8\!\mathcal{N}_k} {}^8\!\phi^{n}_{k} {\widehat{\mathcal{L}}^h_{2,n}},
\end{equation}
where ${}^i\!\mathcal{N}_k$ contains elements $\{n\}$ such that $\left \| {}^i\!\phi^n_{k} \right \|_F \geq {}^i\!\zeta$.

%% file: trajectory_planning.tex
\label{sec:traj-planning}

\subsection{Trajectory Planning Overview}
\label{subsec:planner-overview}

In Section \ref{sec:reachability} we presented our method to predict state uncertainty along a single nominal trajectory. The objective of the trajectory planner is to use this method to explore the environment and find an optimal trajectory for the problem defined in Equation (\ref{eqn:objective}). In this paper we choose the planning framework presented in \cite{ichter2017real} which is desirable given its highly parallelizable structure. However, note that our method to predict state uncertainty can be used with other planning frameworks that consider uncertainty \cite{bry2011rapidly,costante2016perception}.

\begin{figure}[t]
    \centering
  \subfloat{%
       \includegraphics[width=\linewidth,trim={0 0 0.49cm 0},clip]{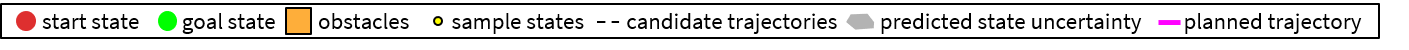}}
    \hfill
  \vspace{-0.2cm}
  \addtocounter{subfigure}{-1}
  \subfloat[]{%
       \includegraphics[width=0.24\linewidth]{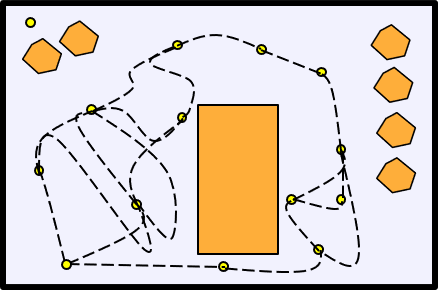}}
    \hfill
  \subfloat[]{%
       \includegraphics[width=0.24\linewidth]{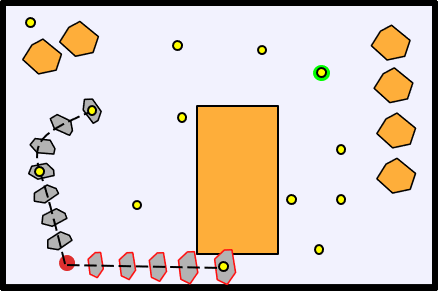}}
    \hfill
  \subfloat[]{%
       \includegraphics[width=0.24\linewidth]{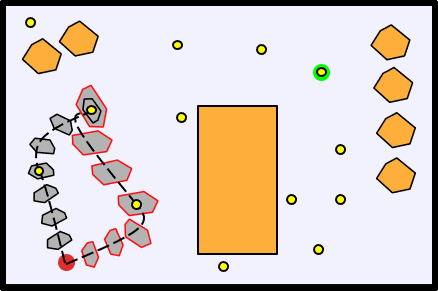}}
    \hfill
  \subfloat[]{%
       \includegraphics[width=0.24\linewidth]{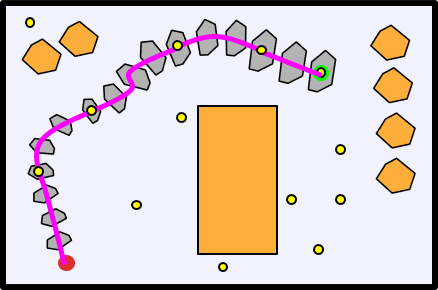}}
    \hfill
  \caption{Overview of trajectory planning framework \cite{ichter2017real}. (a) The planner first builds a graph of feasible trajectories. Then it explores the graph rejecting (b) collision unsafe and (c) undesirable trajectories. (d) The planner stops exploring once it finds a trajectory to the goal state.}
  \label{fig:trajectory-planner-overview}
\end{figure}

For details of the trajectory planning framework, we refer the readers to \cite{ichter2017real}. Here we illustrate the framework in Fig. \ref{fig:trajectory-planner-overview} and summarize it as follows:
\begin{enumerate}
    \item The planner begins with an offline graph constructing phase. Multiple states are sampled in the environment and a graph of kinematically feasible and collision-free trajectories is constructed.
    \item Next, the planner explores the graph by predicting state uncertainty along candidate trajectories. Intersection between the predicted state uncertainty and the obstacles is checked in order to maintain the desired level of collision safety as specified in Equation (\ref{eqn:objective}).
    \item For computational tractability, the planner removes undesirable trajectories by comparing candidate trajectories that arrive at the same state. Here the comparison is done in terms of the trajectory lengths and the sizes of the predicted state uncertainties.
    \item Once a collision-safe trajectory to the goal state is found, the planner stops the exploration phase.
\end{enumerate}
Thus, in order to use our method of predicting state uncertainty along with the above planning framework, we need to design a metric to represent the size of the predicted state uncertainty.

\subsection{Metric for Predicted State Uncertainty}

The state uncertainty along a candidate trajectory is predicted using Equation (\ref{eqn:predicted-state-uncertainty}). Given that we use probabilistic zonotopes as our set representation, the state uncertainty can be expressed in the form of Equation (\ref{eqn:probabilistic-zonotope}) as:
\begin{equation}
\label{eqn:state-uncertainty}
    \underline{\mathcal{X}}_k = {\mathscr{Z}}(c_{\underline{\mathcal{X}}_k}, G_{\underline{\mathcal{X}}_k}, \Sigma_{\underline{\mathcal{X}}_k}),
\end{equation}
where $c_{\underline{\mathcal{X}}_k}$ is the center of the probabilistic zonotope, $G_{\underline{\mathcal{X}}_k}$ is the generator matrix for the bounded uncertainty, and $\Sigma_{\underline{\mathcal{X}}_k}$ is the covariance for the stochastic uncertainty.

\begin{figure}[b]
\centering
  \includegraphics[width=0.5\linewidth,trim={0 0cm 0 0.5cm},clip]{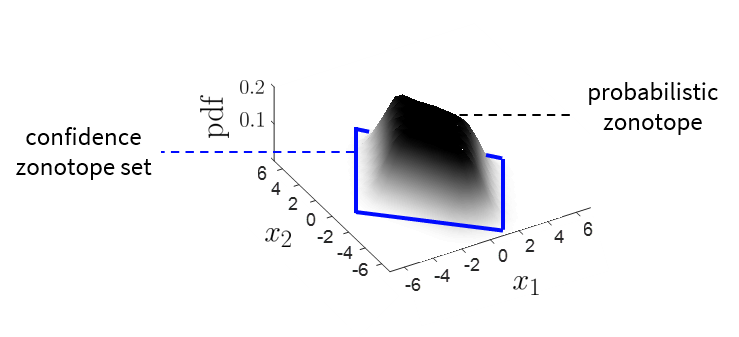}
  \caption{Confidence zonotope of a probabilistic zonotope.}
  \label{fig:confidence-set}
\end{figure}

In order to obtain a metric for the size of a probabilistic zonotope, we perform the following steps:
\begin{enumerate}
    \item Generate a confidence zonotope for the state uncertainty as shown in Fig. \ref{fig:confidence-set}. We represent the confidence zonotope as $\mathcal{Z}(c_{\text{conf}}, G_{\text{conf}})$, with center $c_{\text{conf}} = c_{\underline{\mathcal{X}}_k}$ and generator matrix:
    \begin{equation}
        \label{eqn:concatenated-G-matrix}
    G_{\text{conf}} = [G_{\underline{\mathcal{X}}_k}, \alpha\sqrt{\lambda_1}v_1, \cdots, \alpha\sqrt{\lambda_n}v_n ],
    \end{equation}
    where $\alpha$ is a scalar that follows a chi-square distribution \cite{wang2015confidence,hoover1984algorithms} based on the collision probability threshold from Equation (\ref{eqn:objective}), $\lambda$ and $v$ represent the eigenvalues and eigenvectors of the covariance matrix $\Sigma_{\underline{\mathcal{X}}_k}$, and $n$ here is the dimension of the system.
    \item Use the covariation of the confidence zonotope as the metric for the size of $\underline{\mathcal{X}}_k$. The covariation of a zonotope is defined as $\text{trace}(G^\top_{\text{conf}}G_{\text{conf}})$ \cite{combastel2015zonotopes}.
\end{enumerate}

%% file: simulations.tex
\label{sec:simulations}

\begin{figure}[t]
    \centering
  \subfloat[]{%
       \includegraphics[width=0.35\linewidth]{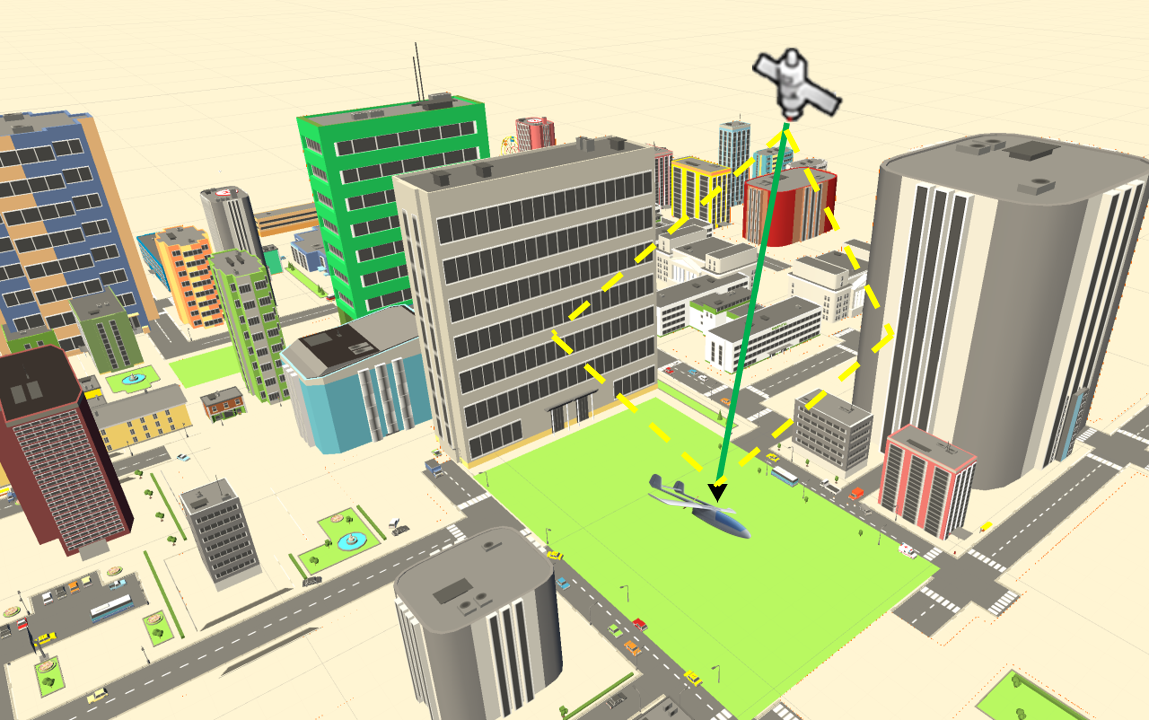}}
    \hspace{2cm}
  \subfloat[]{%
        \includegraphics[width=0.35\linewidth]{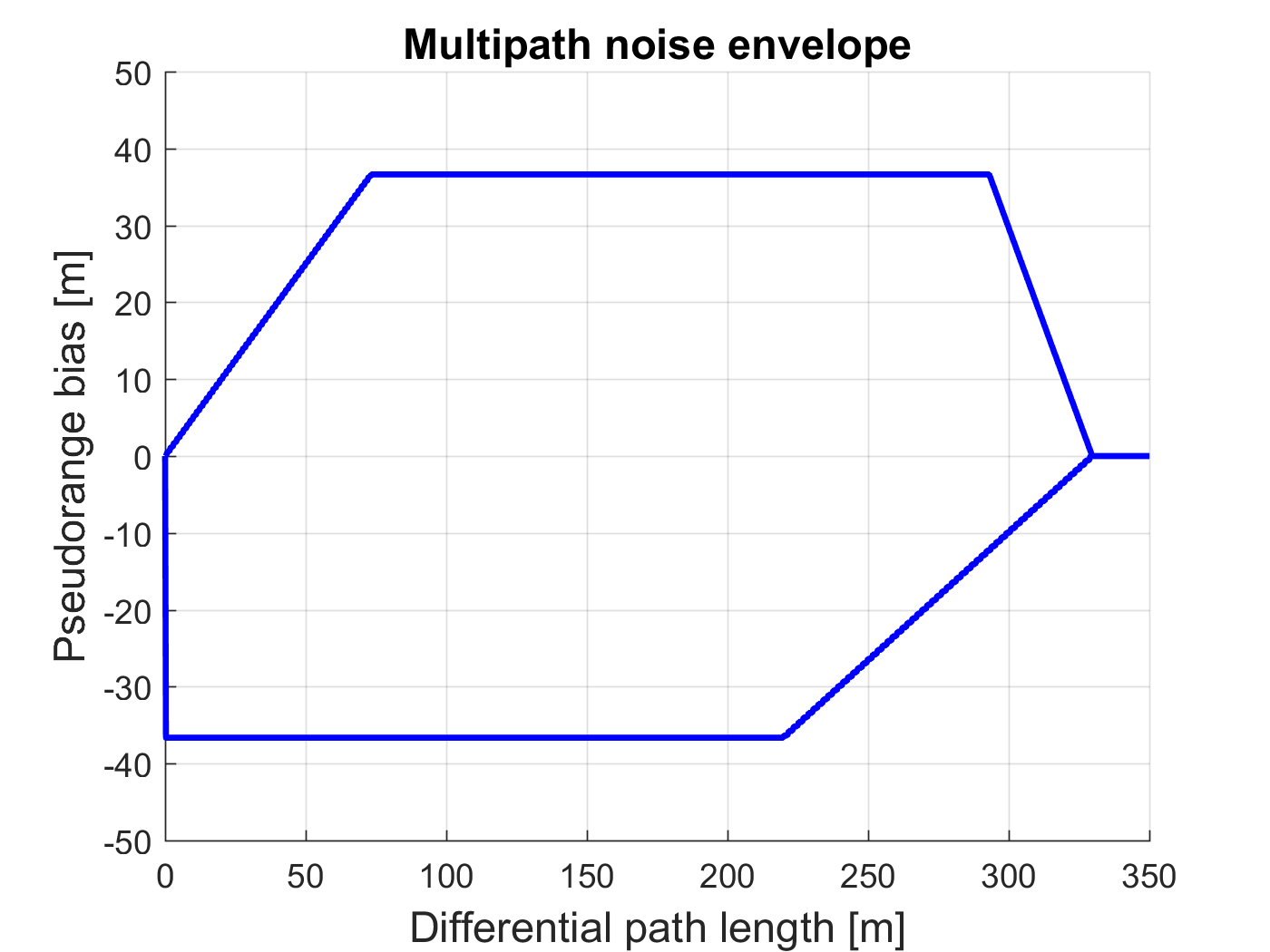}}
    \hfill
  \caption{(a) Simulated urban environment where we perform ray-tracing to compute the differential path lengths between the direct signal (green) and the multipath signals (yellow). (b) Multipath noise envelope for a receiver with a quarter-chip early/late correlator spacing.}
  \label{fig:sim-env-multipath}
\end{figure}

In this section we demonstrate the applicability of the trajectory planner for GNSS-based navigation of fixed-wing UAS in urban environments. We first describe the motion and sensing models we use for the simulations, and then we discuss trajectory planning results for two scenarios: single UAS in a static environment and multiple UAS in a shared airspace.

\subsection{Motion and Sensing Models}

For simplicity we restrict the fixed-wing UAS motion to a horizontal plane and represent it by a 2D Dubins model as follows:
\begin{equation}
    \label{eqn:fixed-wing}
    \underbrace{\begin{bmatrix} {x_1}_k\\ {x_2}_k\\ {\theta}_k \end{bmatrix}}_{x_k} = \underbrace{\begin{bmatrix} {x_1}_{k-1}\\ {x_2}_{k-1}\\ {\theta}_{k-1} \end{bmatrix} + \begin{bmatrix} V_{k-1}\cos(\theta_{k-1})\Delta t\\ V_{k-1}\sin(\theta_{k-1})\Delta t\\ \omega_{k-1}\Delta t\end{bmatrix}}_{f(x_{k-1},u_{k-1})} + w_k,
\end{equation}
where the state vector consists of the 2D position $(x_1, x_2)$ and the heading angle $\theta$, the inputs to the system are the forward velocity $V$ and the angular velocity $\omega$, and $\Delta t$ is the time-step that we set as \SI{0.2}{\second}. We set the covariance matrix for motion model error $w_k$ as $Q = \text{diag}([\SI{0.01}{\square\metre}\ \SI{0.01}{\square\metre}\ \SI{0.001}{\square\radian}])$.

For the sensing model, we consider GNSS pseudorange measurements and heading measurements from an on-board compass. The GNSS pseudorange measurement from the $i^{th}$ satellite can be expressed as \cite{misra2006global}:
\begin{equation}
    \label{eqn:pseudorange}
    \rho^{(i)}_{k} = r(x_k,x^{s_i}_k) + c\delta t + \nu_k^{(i)},
\end{equation}
where $r$ represents the true range between the receiver position $x_k$ and satellite position $x^{s_i}_k$ (which we simulate from publicly available almanac data), $c\delta t$ is the clock bias error, and $\nu^{(i)}_k \sim \mathcal{N}(b_k^{(i)},R^{(i)}_k)$. Here $\nu^{(i)}_k$ contains both the bounded uncertainty due to additional multipath bias $b^{(i)}_k$ and the stochastic uncertainty with covariance $R^{(i)}_k$. We model the stochastic uncertainty with an elevation-based factor \cite{tay2013weighting,hartinger1999variances} as: $R^{(i)}_k = \Sigma_{\rho} / \sin^2(\text{el}^{(i)})$, where we set $\Sigma_{\rho} = \SI{5}{\square\metre}$. Since we are primarily concerned with the UAS position states, we assume for simplicity that the receiver clock and the satellite clocks are perfectly synced, i.e. there is zero clock bias error. However, if desired, clock bias states can also be included in the state vector for the trajectory planner. Thus, given $N$ GNSS satellites, the measurement model for the fixed-wing UAS looks as follows:
\begin{equation}
    \label{eqn:sim-measurement-model}
\underbrace{\begin{bmatrix} z^{(1)}_k\\ \vdots \\ z^{(N)}_k\\ z^{(N+1)}_k \end{bmatrix}}_{z_k} = \underbrace{\begin{bmatrix} r(x_k,x^{s_1}_k) \\ \vdots \\ r(x_k,x^{s_N}_k) \\ \theta_k \end{bmatrix}}_{h(x_k)} + \underbrace{\begin{bmatrix} \nu^{(1)}_k \\ \vdots \\ \nu^{(N)}_k \\ \nu^{(N+1)}_k \end{bmatrix}}_{\nu_k},
\end{equation}
where $z^{(i)}_k = \rho^{(i)}_k \ \forall \ i = 1$ to $N$, and $z^{(N+1)}_k$ represents the heading measurement with $\nu^{(N+1)}_k \sim \mathcal{N}(0,\SI{0.001}{\square\radian})$. For the multipath bias in the pseudorange measurements, we assume a quarter-chip spacing between the early and late correlators \cite{misra2006global}. We perform ray-tracing as shown in Fig. \ref{fig:sim-env-multipath}(a) and assume that the reflected signal strength can be as strong as the direct LOS signal. Based on these characteristics, we get a multipath noise envelope as shown in Fig. \ref{fig:sim-env-multipath}(b).

\subsection{Trajectory Planning Results}
\begin{figure}[t]
    \centering
  \subfloat[]{%
       \includegraphics[width=0.37\linewidth]{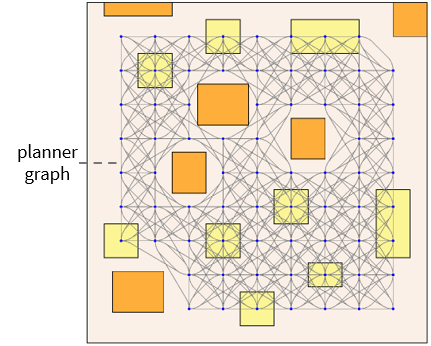}}
    \hfill
  \subfloat[]{%
       \includegraphics[width=0.5\linewidth,trim={0cm 0cm 0cm 0cm},clip]{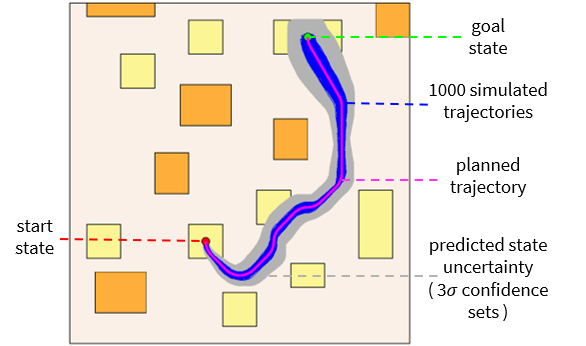}}
    \hfill
  \caption{Planning results for a single fixed-wing UAS. (a) Graph of feasible trajectories built by planner. (b) All $1000$ simulated trajectories lie inside the predicted state uncertainty confidence sets, thus validating collision safety of the planned trajectory.}
  \label{fig:results-single-UAS}
\end{figure}
Given the above motion and sensing models, we setup a \SI{1}{\kilo\metre} $\times$ \SI{1}{\kilo\metre} wide urban environment to evaluate our trajectory planner. In order to simulate multipath effects, we set the flight altitude to \SI{65}{\metre} and include buildings up to \SI{120}{\metre} tall. We first consider trajectory planning for a single UAS and show the results in Fig. \ref{fig:results-single-UAS}. Buildings taller than the flight altitude are colored orange, whereas buildings shorter than the flight altitude are colored yellow. For constructing the planner graph of kinematically feasible and collision-free trajectories, we use a grid of states with \SI{100}{\metre} spacing as shown in Fig. \ref{fig:results-single-UAS}(a). Once the graph is constructed, the planner explores the graph to find an optimal trajectory between a given start and goal state. In order to statistically validate the collision safety of the planned trajectory, we simulate $1000$ trajectories along the planned trajectory. We observe that all $1000$ simulated trajectories remain within the $3\sigma$ confidence zonotopes of the predicted state uncertainty as shown in Fig. \ref{fig:results-single-UAS}(b). Thus, the trajectory planner safely leads the UAS to the goal state under stochastic and bounded GNSS uncertainties.

\begin{figure}[t]
    \centering
  \subfloat{%
    \hspace{1.7cm}   \includegraphics[width=0.8\linewidth,trim={0 0 0.49cm 0},clip]{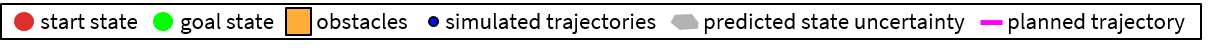}}
    \hfill
  \vspace{-0.2cm}
  \addtocounter{subfigure}{-1}
  \subfloat[]{%
       \includegraphics[width=0.3\linewidth,trim={0 -0.6cm 0 0},clip]{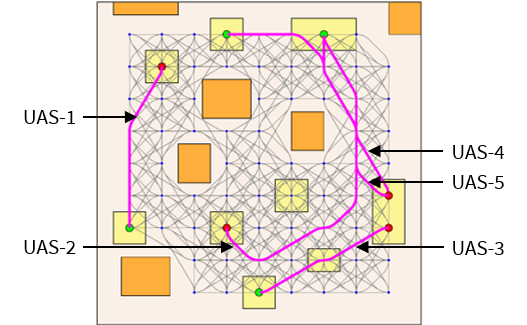}}
    \hfill
  \subfloat[]{%
       \includegraphics[width=0.3\linewidth,trim={5.8cm 4cm 4cm 4cm},clip]{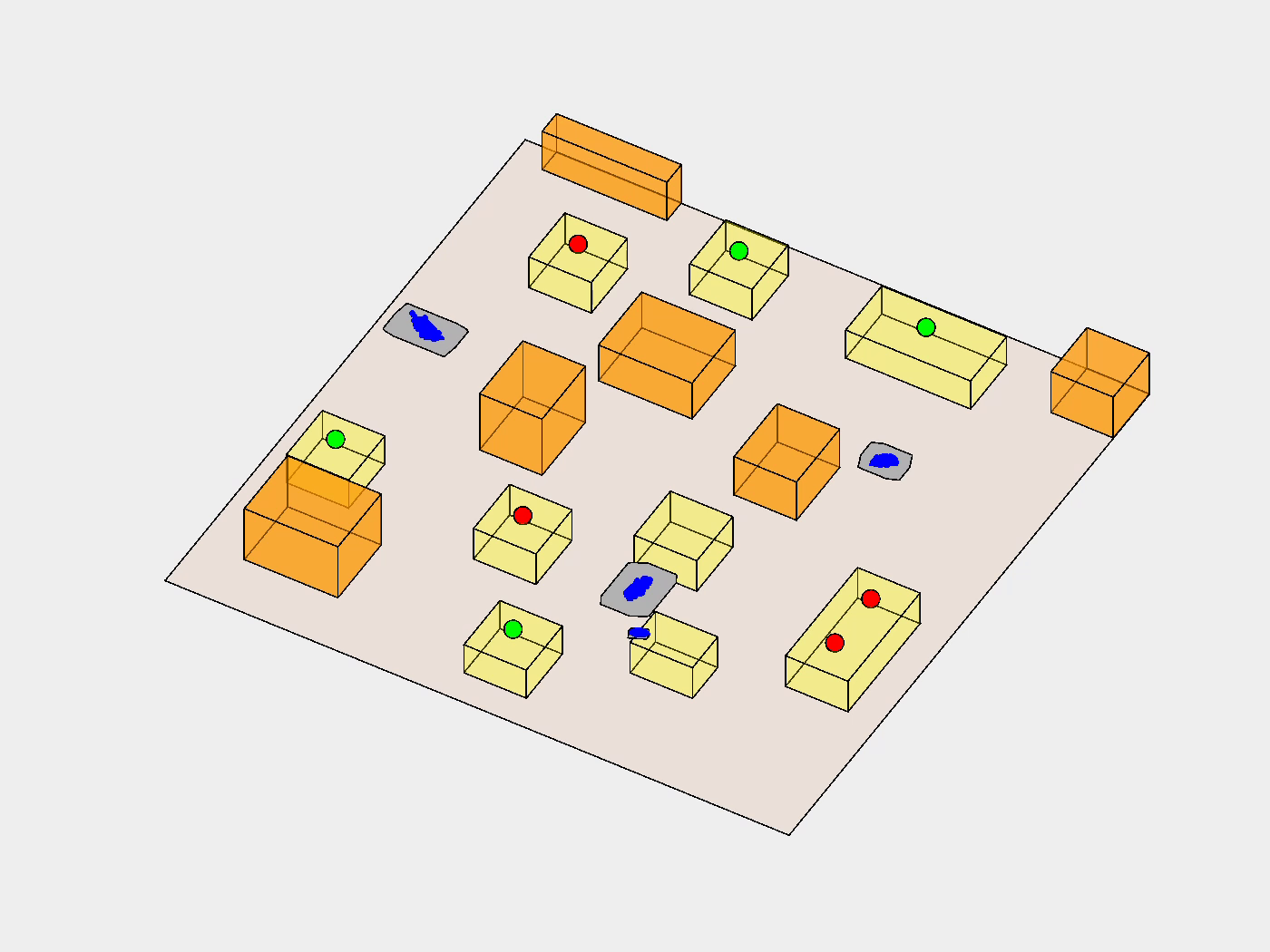}}
    \hfill
  \subfloat[]{%
       \includegraphics[width=0.3\linewidth,trim={5.8cm 4cm 4cm 4cm},clip]{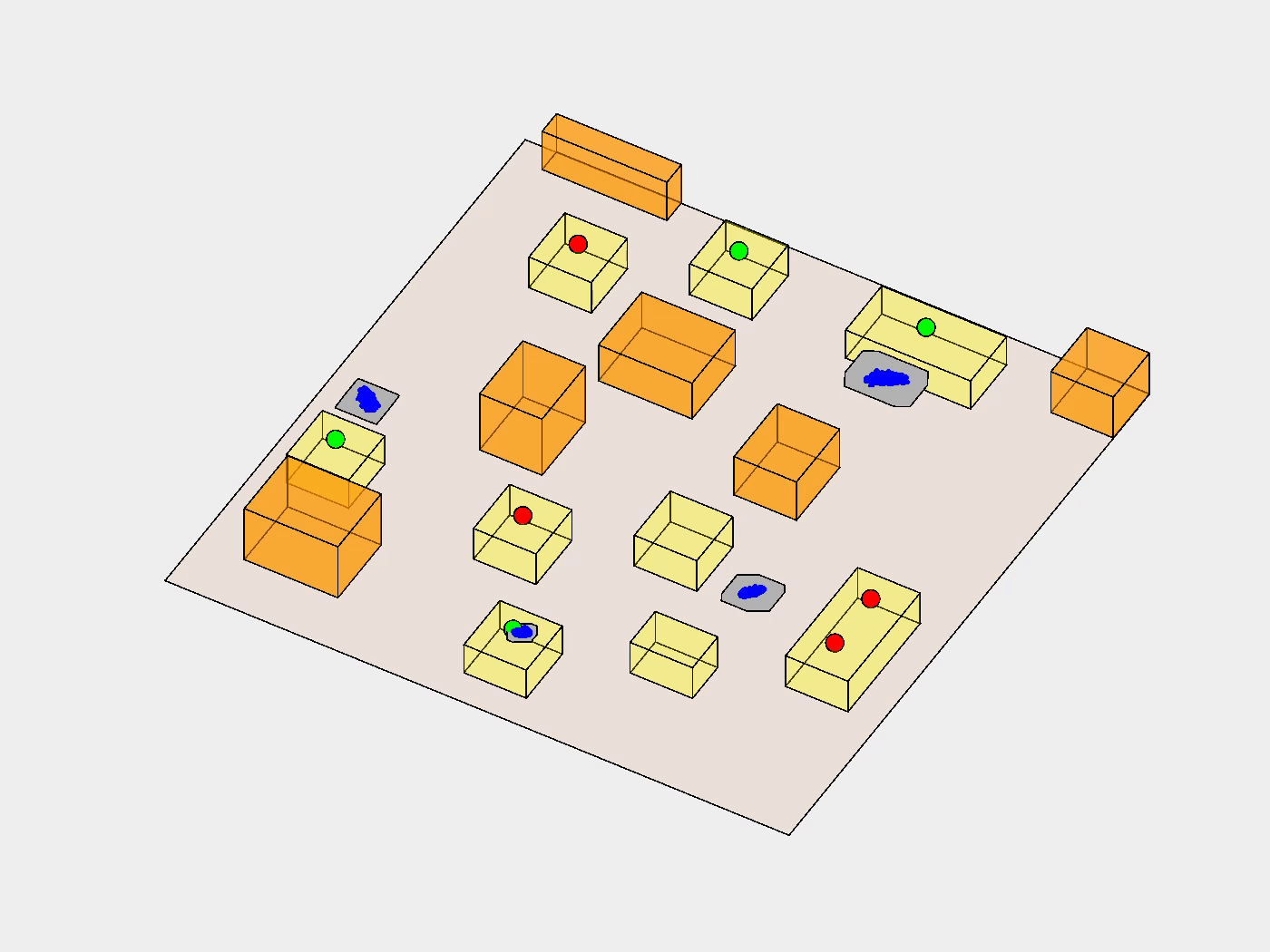}}
    \hfill
  \subfloat[]{%
       \includegraphics[width=0.3\linewidth,trim={5.8cm 4cm 4cm 4cm},clip]{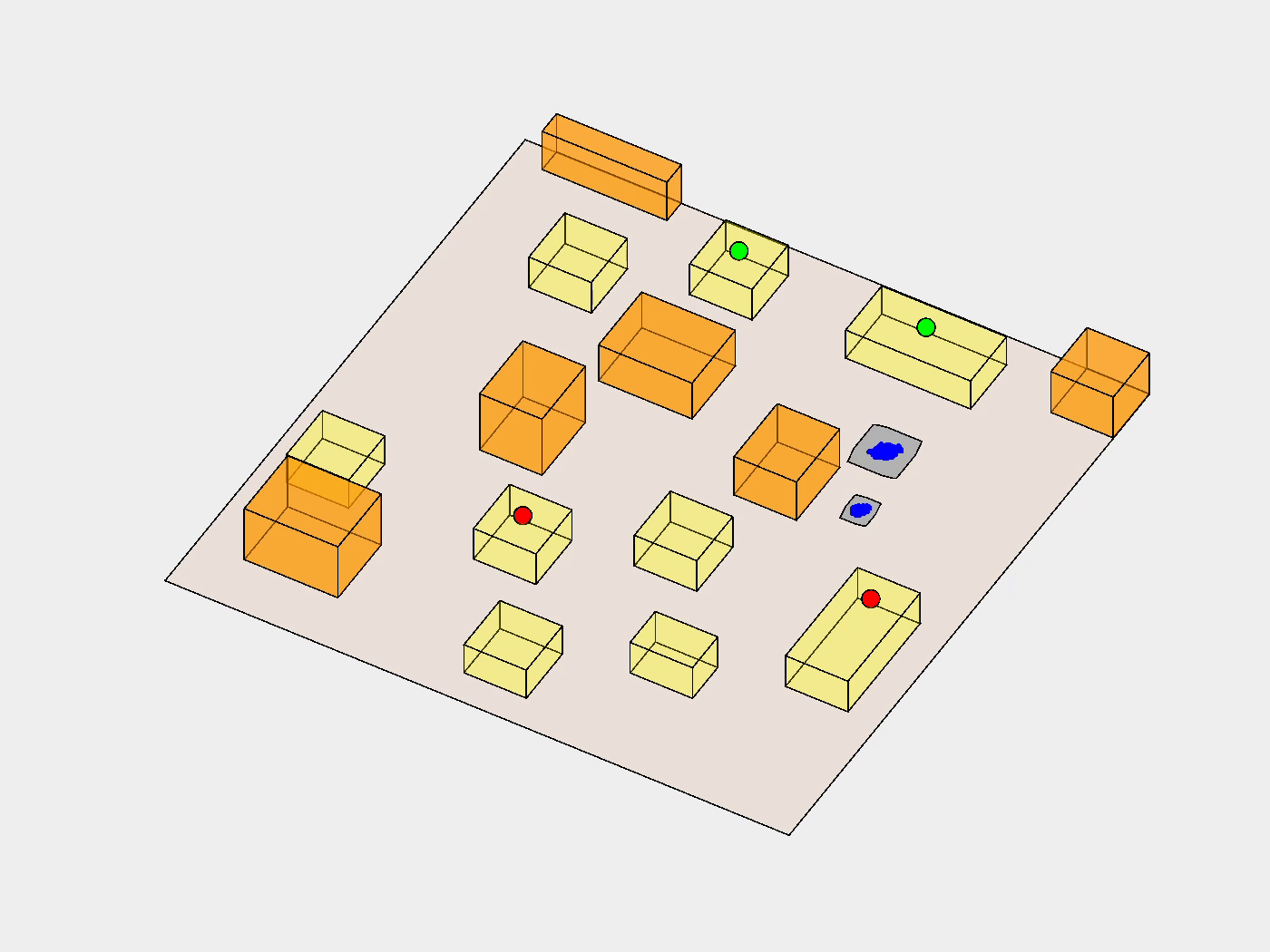}}
    \hfill
  \subfloat[]{%
       \includegraphics[width=0.3\linewidth,trim={5.8cm 4cm 4cm 4cm},clip]{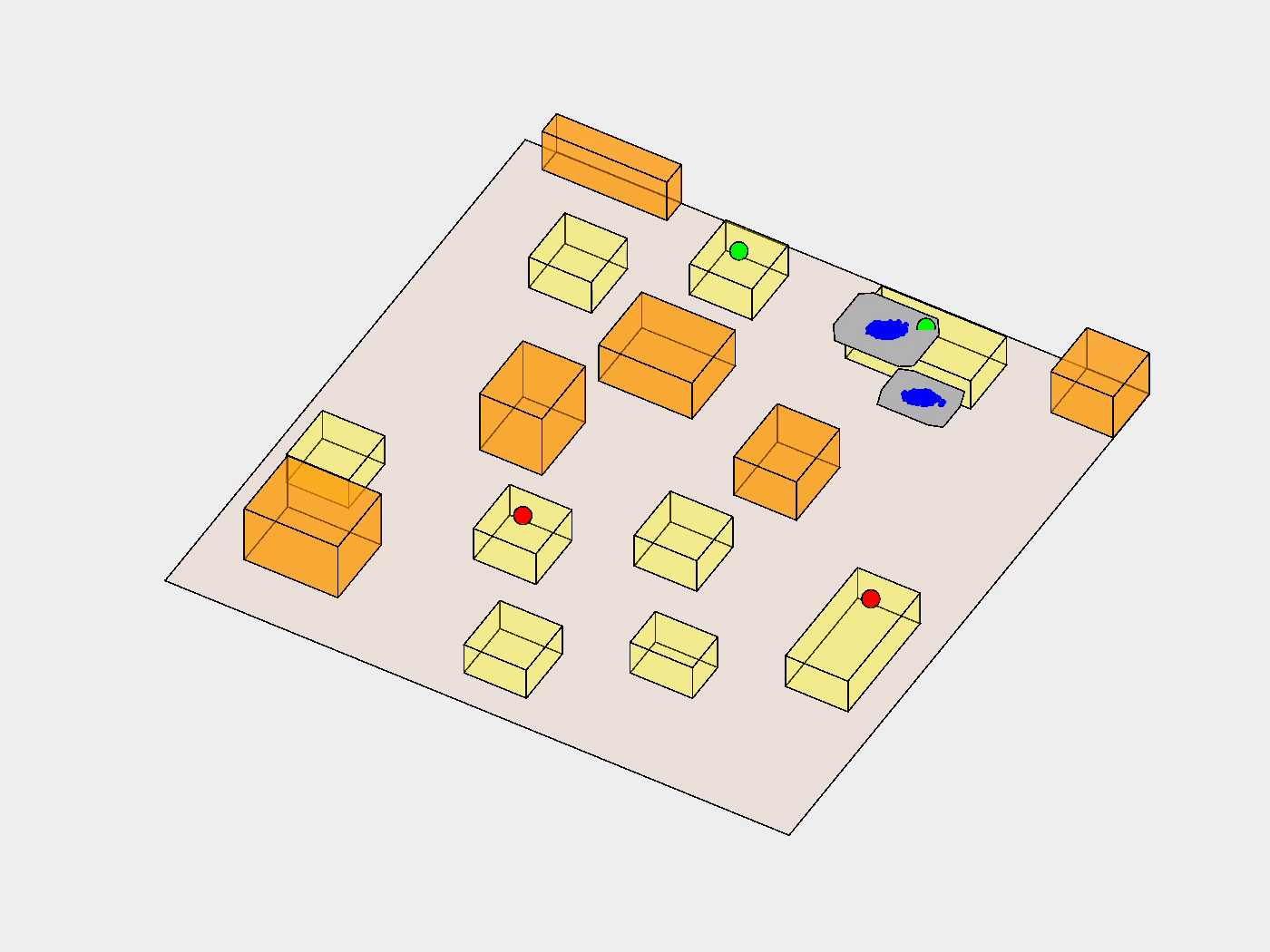}}
    \hfill
  \subfloat[]{%
       \includegraphics[width=0.3\linewidth,trim={5.8cm 4cm 4cm 4cm},clip]{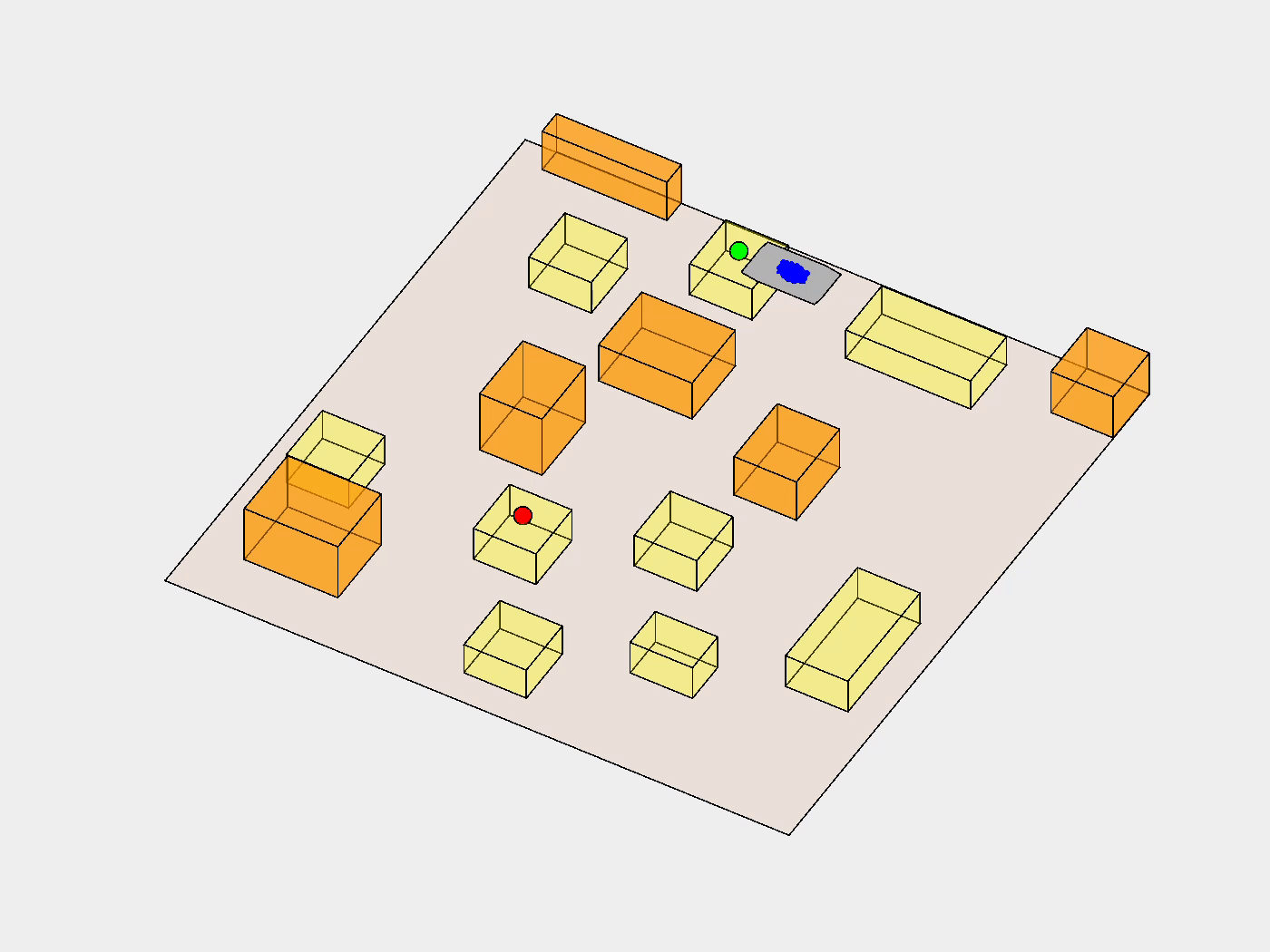}}
    \hfill
  \caption{Planning results for multiple fixed-wing UAS. (a) Planned trajectories for all UAS along with the planner graph. (b)-(f) Snapshots during trajectory execution including $1000$ simulated trajectories and predicted state uncertainty confidence sets. The complete video can be found at https://youtu.be/eyA3vEojdnQ.}
  \label{fig:results-multiple-UAS}
\end{figure}

Next, we evaluate the trajectory planner for multiple UAS in a shared airspace. Here we sequentially plan trajectories for five UAS in the same urban environment. Since we assume all UAS to be fixed-wing with the same motion model in Equation (\ref{eqn:fixed-wing}), the planning framework allows us to reuse the same planner graph constructed in the above scenario. Fig. \ref{fig:results-multiple-UAS}(a) shows the planner graph along with the planned trajectories for the five UAS. The planner begins by finding an optimal trajectory for UAS-1 while maintaining collision safety with respect to the obstacles (orange-colored buildings). Next, the planner finds an optimal trajectory for UAS-2 while maintaining collision safety with respect to UAS-1 and the obstacles. Note that while a shorter candidate trajectory existed for UAS-2, it was rejected by the planner as it could not maintain collision safety with respect to UAS-1. Next, the planner finds an optimal trajectory for UAS-3 while maintaining collision safety with respect to UAS-1, UAS-2, and the obstacles. For UAS-4 the planner finds an optimal trajectory that seems to collide with the trajectory for UAS-2. However, the trajectories do not intersect temporally, thus maintaining collision safety. For UAS-5, we set the same start state as for UAS-4 and set the trajectory start time a few seconds after that of UAS-4. Again the optimal trajectory found by the planner does not intersect temporally with the trajectories for UAS-2 and UAS-4, thus maintaining collision safety. In order to statistically validate the collision safety of the planned trajectories for the five UAS, we simulate $1000$ trajectories for each UAS. Figs. \ref{fig:results-multiple-UAS}(b)-(f) show snapshots of the UAS flights with the simulated trajectories and the $3\sigma$ confidence zonotopes for the predicted state uncertainties. All $1000$ simulated trajectories for all the five UAS remain within the $3\sigma$ confidence zonotopes, thus demonstrating the applicability of the planner to plan collision safe UAS trajectories in a shared airspace.

%% file: conclusion_future.tex
\label{sec:conclusion_future}

We have presented a trajectory planner that maintains collision safety in the presence of stochastic and bounded sensing uncertainties. We first improved our prior reachability analysis by computing reachable sets as a function of independent quantities. The expression for computing the reachable set was then approximated to predict state uncertainty along the trajectory. Next, we combined our method to predict state uncertainty with a highly parallel trajectory planning framework. We designed a metric for the size of the state uncertainty by obtaining the covariation of its confidence zonotopes. Finally, we demonstrated the applicability of the trajectory planner for GNSS-based navigation of fixed-wing UAS in urban environments. We considered multiple scenarios and statistically validated the collision safety of the planned trajectories.